\documentclass[sigconf]{acmart}
\usepackage{tabularx}
\usepackage{subcaption}
\usepackage{multirow}
\usepackage{booktabs}

\copyrightyear{2025}
\acmYear{2025}
\setcopyright{cc}
\setcctype{by}
\acmConference[FDG '25]{International Conference on the Foundations of Digital Games}{April 15--18, 2025}{Graz, Austria}
\acmBooktitle{International Conference on the Foundations of Digital Games (FDG '25), April 15--18, 2025, Graz, Austria}
\acmPrice{}
\acmDOI{10.1145/3723498.3723820}
\acmISBN{/25/04}

\begin{document}

\title{Analysis of Robustness of a Large Game Corpus}

\author{Mahsa Bazzaz}
\affiliation{%
  \institution{Northeastern University}
  \city{Boston, Massachusetts}
  \country{USA}}
\email{bazzaz.ma@northeastern.edu}
\orcid{0009-0004-0022-9611}

\author{Seth Cooper}
\affiliation{%
  \institution{Northeastern University}
  \city{Boston, Massachusetts}
  \country{USA}}
\email{se.cooper@northeastern.edu}
\orcid{0000-0003-4504-0877}

\begin{abstract}
Procedural content generation via machine learning (PCGML) in games involves using machine learning techniques to create game content such as maps and levels. 2D tile-based game levels have consistently served as a standard dataset for PCGML because they are a simplified version of game levels while maintaining the specific constraints typical of games, such as being solvable. In this work, we highlight the unique characteristics of game levels, including their structured discrete data nature, the local and global constraints inherent in the games, and the sensitivity of the game levels to small changes in input. We define the robustness of data as a measure of sensitivity to small changes in input that cause a change in output, and we use this measure to analyze and compare these levels to state-of-the-art machine learning datasets, showcasing the subtle differences in their nature. We also constructed a large dataset from four games inspired by popular classic tile-based games that showcase these characteristics and address the challenge of sparse data in PCGML by providing a significantly larger dataset than those currently available.
\end{abstract}
\maketitle

\section{Introduction} \label{sections:introduction}
``Bad data'' often arises when real-world information fails to conform to predefined rules or constraints, causing breakdowns in how that information is processed and interpreted~\citep{mccallum2012bad}. In the case of highly structured discrete data, these constraints can be especially unforgiving: even a minor deviation from expected formats or delimiters can cause parsing failures or lead to incorrect assumptions about the underlying content. Examples of these structured discrete data include many programming languages like Python or many file formats like CSV where substituting a comma for a semicolon (or vice versa) may cause an entire data file to fail to parse.

Game levels are also highly structured discrete data in which a ``tile'' is the smallest building block that constructs the visual and functional layout of a game's environment, typically corresponding to specific gameplay mechanics or aesthetics (e.g., walls, enemy). In contrast, a pixel in an image is the smallest visual element that composes the display of the image, defining its resolution and color detail without inherent functionality beyond contributing to the overall picture. The main difference lies in their functionality: while changing a single pixel will not typically impact the output due to the large number of pixels and their redundancy, altering a single tile can notably affect gameplay or the visual aesthetics.

The term ``hard constraint'' originates from the field of constraint satisfaction.  Constraint satisfaction problems involves finding assignments for variables that satisfy specific constraints on those variables~\citep{smith2012case}. Constraints in this context can be categorized as either hard or soft. Hard constraints impose mandatory conditions that the variables must satisfy. In contrast, soft constraints allow some flexibility; if the conditions on the variables are not fully met, the objective function includes penalties proportional to the degree of the violation~\citep{EDELKAMP2012571}.

When discussing hard \emph{global constraints} in game levels, \emph{solvability} is a major concern. A solvable game level is one which is possible for the player to complete, e.g. the player can move from the start to the end position through a series of valid movement patterns, potentially completing optional tasks and avoiding obstacles. This is a global constraint that must be satisfied when generating game levels. Other global constraints can arise from the tasks that the player needs to complete during gameplay. These may include having keys and corresponding doors or treasures, or incorporating a specific number of enemies or obstacles at varying difficulty levels. 

\begin{figure*}[ht]
    \includegraphics[width=\textwidth]{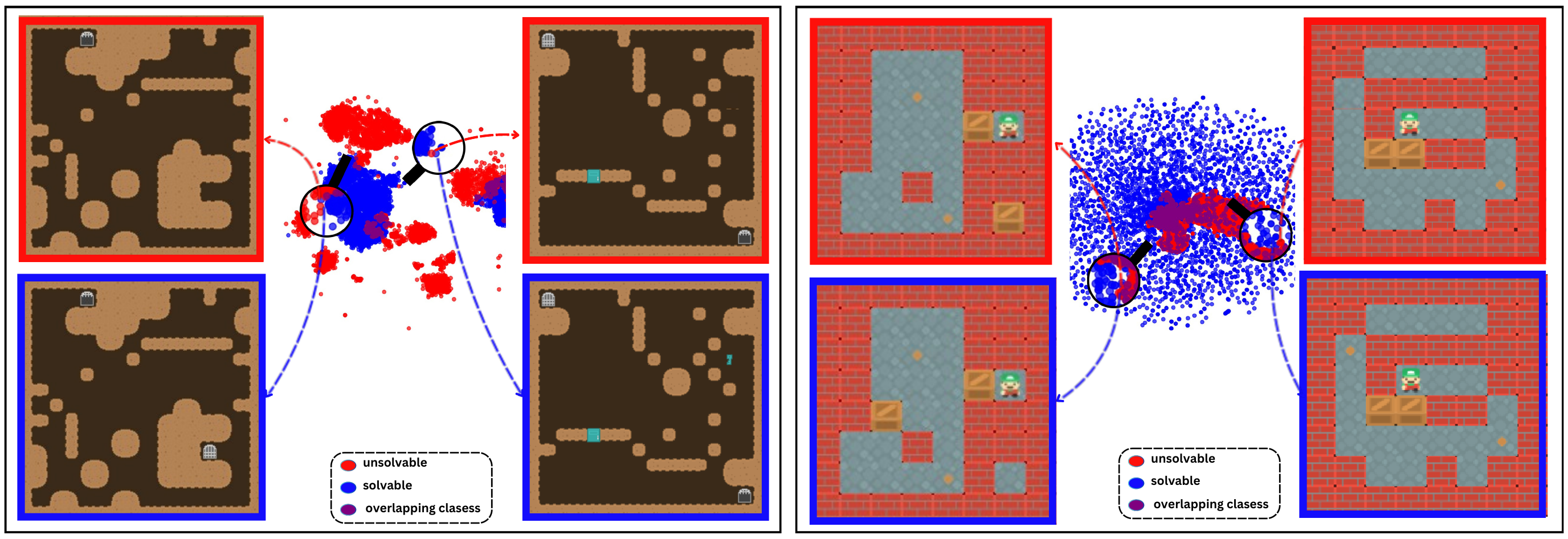}
    \caption{Solvable (blue) and unsolvable (red) levels can be only different by a single tile change or swap. Unacceptable levels in the upper row become the acceptable level in the lower row after changing or swapping a single tile.}
    \label{fig:change}
\end{figure*}

\begin{figure}[h]
    \includegraphics[width=0.45\textwidth]{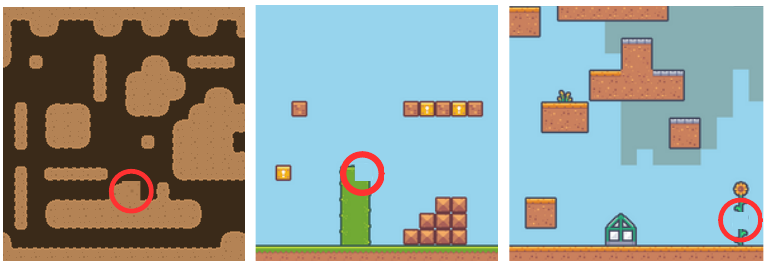}
    \caption{Sample examples of violation of local constraints in different games. (From left to right: \texttt{Cave)}, \texttt{Platform}, and \texttt{Vertical}}
    \label{fig:broken}
\end{figure}

In addition to global constraints, there are also local constraints in game levels that need to be satisfied to produce \emph{acceptable} levels. The most commonly overlooked local constraints in procedurally generated levels are often local structures that end up with pieces missing or arranged incorrectly.  Such structures include pipes in platformer games and broken decorations in puzzle games. solvable and acceptable levels can be directly used in the corresponding game while unsolvable or unacceptable levels cannot.
These levels may become solvable and acceptable after a repair process to enforce the satisfaction of these constraints manually. Figure \ref{fig:broken} shows what a typical broken structure can look like in our sample games.

A level may be solvable with satisfied global constraints, but not acceptable if local constraints are not met. Enforcing these local constraints is not a trivial task \citep{gonzalez2022mario}, so much so that even in good-performing state-of-art models we see \emph{unacceptable} generated levels with some sort of violation of local constraints (e.g. transformer based \citep{sudhakaran2024mariogpt}, VAE based \citep{sarkar2020sequential}, and GAN based models \citep{volz2018evolving}).

Satisfying the global hard constraint of solvability in games can be very challenging because of a very specific attribute of some game levels. Unlike popular datasets like CIFAR-10 or MNIST, where altering a single pixel does not fundamentally alter the image's identity (e.g., turning a dog into a cat), the alteration of a single tile in a dataset of game levels can drastically impact solvability, making an entire level unsolvable. For instance, in the work by \citet{sorochan2021generating} on generating Lode Runner levels using LSTMs, some generated levels were unacceptable just because of the placement of enemy tiles directly beside the player's starting tile in the game (leaving no viable way for the player to avoid or trap the enemy).

In addition, the inherited discrete structure in games proposes challenges due to the inability to back-propagate during the training of Neural Networks. Such challenges are dealt with by continuous approximation methods of discrete structures \citep{jang2016categorical}. However, in some applications like game levels with hard constraints, one cannot replace a hard constraint with a soft constraint.

In particular, designers encounter many situations where levels must satisfy multiple global constraints besides solvability. They need models capable of creating controllable content, such as generating easier or harder levels, or adding specific quests. In these cases, a tile can represent a hard constraint in the generation process, like the existence of a hidden key essential for solving a puzzle. When dealing with hard constraints, these elements cannot be ignored or replaced with soft constraints. In such cases when the number of hard constants adds up, such tiles can be ignored through the soft constraint counterpart. 

All the points mentioned above, cause training reliable models in these tasks to be much harder, which results in work on PCGML often involving a post hoc cleaning step \citep{khalifa2016general,MawhorterProcedural,6374170}. This makes such content generation not scalable making the task of generating high-quality scalable game level a challenge. 


Many communities, including but not limited to game developers, work on generative models that require a form of \textbf{correctness} \citep{kusner2017grammar}. In this work, we aim to formalize this issue as a step toward bridging these communities. We define the concept of robustness to quantify the sensitivity of game levels to input changes by assessing the likelihood that very similar levels share the same label within a distribution. Additionally, we compare the robustness of game levels to that of state-of-the-art machine learning datasets, highlighting the key differences.

Inspired by the VGLC \citep{VGLC}, this research also introduces an expansive corpus comprising various 2D tile-based games. This dataset specifically highlights the robustness of game level data, and expands the pool available for PCGML research. Also, we believe that other communities working with structured data requiring correctness may also find value in our dataset, as the underlying tasks share significant similarities. Moreover, game-level generation can be conducted entirely in silicon, thereby mitigating potential ethical concerns associated with other data generation methods.

Unlike existing game levels, the levels in this corpus have been specifically generated for custom games across various genres.  Thus we call it the Generated Game Level Corpus (GGLC). Each game has thousands of levels, greatly expanding the scale of the dataset. The dataset includes game levels of different sizes, aesthetics and looks, and difficulties, with different local and global constraints. The GGLC, along with setup documentation, is publicly available on GitHub\footnote{\url{https://github.com/TheGGLC}}.


In summary, our contributions are as follows:
\begin{itemize}
  \item We investigate the phenomena of small changes in the input of game levels causing noticeable changes in the output (resulting in unsolvable or unacceptable levels).
  \item We formalize the problem by introducing the robustness of data as a measure of this sensitivity to input change, and we compare robustness game levels to popular benchmark datasets.
  \item We introduce a large and diverse dataset of solvable and unsolvable game levels of four different 2D tile-based games coupled with level solutions (path or playthrough) that capture this sensitivity to input changes.
\end{itemize}

\begin{figure*}[h]
    \includegraphics[width=\textwidth]{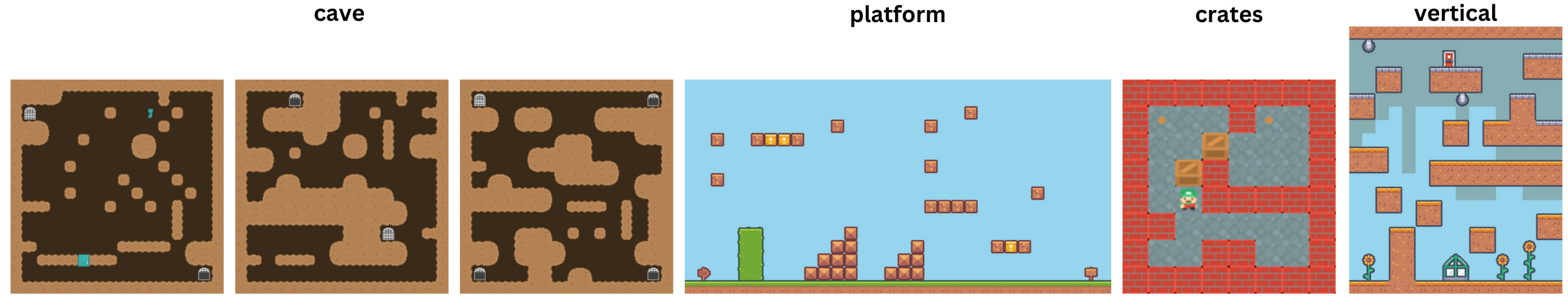}
    \caption{Example sample of each dataset. Cave levels come in different versions.}
    \label{fig:images}
\end{figure*}

\section{Related Work} \label{sections:background}
\subsection{Structured Discrete Data}
Machine learning on structured data is a rapidly growing field that involves significantly more challenges, with many real-world examples that show its unique challenges~\citep{levesque2012winograd}. Various communities are addressing these challenges, particularly for structured discrete data types that require \textbf{correct} outputs, such as arithmetic expressions, source code, molecules, and game levels~\citep{sui2015safe}.

One example is molecules which can be represented using SMILES strings, which are a form of notation that describes molecular structures with short ASCII strings~\citep{weininger1988smiles}. However, one immediate challenge of using strings to represent molecules is that small changes in a string can lead to completely different molecular structures, or result in representations that do not correspond to valid molecules.

A more recent approach, SELFIES, was developed to provide a robust and machine-readable molecular representation~\citep{krenn2020self}. Unlike SMILES, SELFIES addresses the limitations in encoding that can produce invalid molecular structures. It utilizes a self-referencing, grammar-based method, ensuring that any valid SELFIES string will always represent a chemically valid structure.

To further tackle the challenges of generative modeling with structured discrete data, the Grammar Variational Autoencoder (GVAE) encodes and decodes directly from parse trees derived from context-free grammars. This method guarantees that the generated outputs are always valid~\citep{kusner2017grammar}.

Another relevant tool is Poli~\citep{Gonzalez-Duque:poli:2024}, a library designed for creating and managing black-box objective functions, primarily aimed at structured discrete sequence optimization tasks. The name ``Protein Objectives Library'' reflects its applications in fields like drug design and protein engineering, where proteins and small molecules are represented as discrete sequences for optimization. Poli includes examples of various tasks, such as protein optimization, molecule optimization, and even the generation of Super Mario levels.

Creating game levels poses unique challenges compared to working with state-of-the-art machine learning datasets, primarily due to the presence of both global and local hard constraints. Each tile placement must satisfy local aesthetic and structural constraints, as well as global solvability constraints. These complexities can cause state-of-the-art models to struggle.

Research by \citet{sturtevant2020unexpected} has previously explored this phenomenon in games, demonstrating that minor modifications to existing levels can significantly increase the complexity and solution length of puzzles that may not be easily anticipated by human designers. Their quantitative analysis and user study revealed that even small incremental design adjustments can greatly affect the overall experience of playing a level. For instance, Figure \ref{fig:change} illustrates how moving a single tile in a \texttt{crates} game can shift the level from solvable to unsolvable. The results of this prior user study that showcases this phenomenon were a major inspiration for this work.

\subsection{Data in Games}
The majority of research in PCGML has relied on a limited selection of established datasets. Constructing these datasets demands significant human effort, typically involving the extraction of data from various online repositories, often curated by enthusiasts.

Some datasets are specifically curated for further studies. For example, \citet{VGLC} developed the Video Game Level Corpus (VGLC) of 428 manually annotated levels from 12 games, including Doom, Doom 2, Mario, and the initial quest of The Legend of Zelda. Additionally, other datasets have emerged as incidental outcomes of research on generative algorithms. For instance, \citet{Zafar_2019} curated a corpus for Angry Birds Levels\footnote{\url{https://github.com/AdeelZafar123/AngryBirdsDataSet}}, including 100 levels from the original Angry Birds game and 100 levels generated by their baseline algorithm. Furthermore, \citet{boxobanlevels} generated an extensive corpus of Boxobon levels, a puzzle game inspired by Sokoban, generated using various models, such as Deep Reinforcement Learning and tree search. \citet{khalifa2019intentional} employed evolutionary algorithms and quality-diversity algorithms to generate a dataset of Super Mario Bros levels. Similarly, \citet{biemer2021gram} assembled datasets featuring levels from Super Mario Bros., Kid Icarus, and DungeonGrams (a custom roguelike game) using N-gram quality-diversity search. Beyond the academic scope, hobbyists have also contributed to the creation of game level datasets, like datasets for Rush Hour \footnote{\url{https://www.michaelfogleman.com/rush/\#DatabaseDownload}} and Connections \footnote{\url{https://connections.swellgarfo.com/archive}}.


Within the scope of machine learning methodologies, the scarcity of training data in gaming contexts continually amplifies the dilemma of model size selection \citep{karth2019addressing}. Larger models tend to overfit and memorize the sparse training data inherent to these domains \citep{guzdial2022procedural}. This issue is particularly pronounced in recent transformer-based approaches. Notably, \citet{todd2023level}'s experimentation with generating Sokoban levels using Large Language Models highlighted the imperative for a nuanced consideration of dataset size to effectively leverage LLMs for game level generation.

The limited number of available training examples in games has forced previous work to find workarounds. For instance, \citet{maurer2021adversarial} instead of training a GAN model, leveraged an adversarial random forest classifier combined with a greedy hill-climbing search. However, their system failed to produce games of human quality. Their experiments showed that as the generated games more closely resembled human-authored games, the classifier's effectiveness as a fitness function diminished.
Many previous studies have tried solving this issue by focusing on few-shot learning and models that need minimal training data. \citet{bontrager2021learning} introduced Generative Playing Networks (GPN), a framework that combines agent policy learning with environment generation. This method requires minimal data and no domain knowledge but is computationally intensive. \citet{torrado2020bootstrapping} proposed a new bootstrapping training procedure, where the solvable generated levels were added to the training set, aiming to reduce duplication and the generation of unsolvable levels. In more recent work, \citet{Schubert2022TOADGAN} presented the Token-based One-shot Arbitrary Dimension Generative Adversarial Network (TOAD-GAN), which generates token-based video game levels from a single example.

\section{Methodology}

The concept of \emph{robustness} was introduced in the context of classifiers to measure the likelihood that similar points receive the same label from a classifier \citep{pmlr-v119-bhattacharjee20a}. The robustness of a classifier $f$ over $\mathcal{D}$, denoted as $A_r(f,\mathcal{D})$ is the probability that $\forall x, x' \in \mathcal{D}$ s.t. $d(x,x') \leq r$, the classifier will predict the same label \citep{pmlr-v119-bhattacharjee20a}.
\begin{equation}
    A_r(f,\mathcal{D}) = \mathbb{P}_{x\sim \mathcal{D}}[f(x) = f(x') | \forall x',\, d(x,x') \leq r]
\end{equation}

The robustness of a classifier $f$ over a distribution $\mathcal{D}$ captures how consistently the classifier behaves under small perturbations of its inputs. Formally, it is defined as the probability that any two points $x$ and $x'$ drawn from $\mathcal{D}$, which lie within a distance $r$ of each other (as measured by some distance metric $d$), receive the same predicted label from $f$. In other words, if $x$ is changed only slightly (within $r$) to become $x'$, a robust classifier should still classify both points identically. Thus, a high value of $A_r(f,\mathcal{D})$ indicates that the classifier is stable against small perturbations in the input space, reflecting its ability to maintain consistent predictions in regions where inputs are close together.

In our context on the other hand, we are interested in quantifying the robustness of the \emph{data}. To do this, we can use the same method by replacing $f(x)$ by the label of $x$.
\begin{equation}
    \text{D}_r(\mathcal{D}) = \mathbb{P}_{x\sim \mathcal{D}}[\text{Label of $x$} = \text{Label of $x'$} | \forall x',\,d(x,x') \leq r]
\end{equation}

For a given radius $r$, the value of $\text{D}_r(\mathcal{D})$ represents how sensitive the data is to changes in the input.

Similarly, the opposite --- non-robustness --- can be formally defined as:
\begin{equation}
    \text{ND}_r(\mathcal{D}) = \mathbb{P}_{x\sim \mathcal{D}}[\text{Label of $x$} \neq \text{Label of $x'$} | \exists x',\,d(x,x') \leq r]
\end{equation}

and by computing $\text{ND}_r(\mathcal{D})$, we can understand how far from being robust the data is.

It is important here to note the difference between the robustness of data and the robustness of a classifier in adversarial training. The subtlety here is the difference between the \textbf{true label} of data and the \textbf{predicted label} of data when using a classifier. The goal of adversarial training is to develop classifiers that are robust to small perturbations, better reflecting the true nature of the data.

For instance, in the case of a CIFAR-10 panda image, the true label (ground truth) of a panda \textbf{should not} change if a single pixel is altered, as a noisy panda is still a panda in human perception. If a classifier changes its prediction due to such minor modifications, it \textbf{fails} to model the true label. Adversarial training aims to prevent this, ensuring that classifiers remain robust to such small attacks. Thus, the ideal classifier for CIFAR-10 should be robust to small input perturbations because, to a human, an image like a panda should not change its classification simply due to a few pixel changes.

However, in the case of a lock and key style game, the ground truth of a level \textbf{should change} if a key is removed from a solvable level --- solvable levels needing a key become unsolvable without it.
Therefore, if a classifier changes its prediction based on the removal of a key tile, it \textbf{correctly} models the true label of the data. Therefore, the ideal classifier for Cave game should not necessarily be robust to changes of all individual tiles, as some changes can significantly affect the true level's solvability.

\section{Domains} \label{sections:datasets}

This work introduces a dataset of four 2D tile-based games created based on popular classic genres using image tiles from \citet{WEB_kenney}, a large source of free and open-source game assets. Table \ref{table:tiles} shows the different tiles used to create games in this dataset.

We focus on generating levels for custom games instead of existing ones to ensure that the dataset can be freely accessible to everyone. This dataset is provided under a Creative Commons CC-BY 4.0 license and software created as part of the generation and analysis is covered by the standard MIT license.

All game levels in this dataset include both global and local constraints dictated by their game mechanics. The universal constraint across all game levels is solvability. This dataset includes different solvable and unsolvable levels for each game to specifically illustrate these criteria and the associated challenges.

One of the biggest advantages of using a constraint-based generator for dataset creation is its flexibility. This approach allows for future expansion of the dataset with new games and additional levels for existing games.

$\bullet$ \texttt{cave}: A simple top-down cave map first introduced by \citet{cooper2022sturgeon}. Cave levels come in three different versions: simple, doors, and portal. In each level, players navigate maze-like paths to find the exit door leading to the next level. In levels with doors, players must locate keys to unlock these doors and progress. Portal levels feature doors that transport players to other locations within the level, resulting in two independent areas that are not reachable from each other --- thus the path through the level is not contiguous. There are more than $5,000$ simple levels, $20,000$ levels with doors, and $35,000$ levels with portals in this dataset. These levels come in $16*16$ size. 

$\bullet$ \texttt{platform}: A simple platformer inspired by Super Mario Bros. \citep{GAME_mariobros}. There are local structures similar to those in level 1-1 of that game. This platformer game includes various platforms arranged at different heights and distances, requiring the player to walk and jump to reach the end from the starting position while avoiding obstacles and avoiding falling off. There are more than $10,000$ levels in this dataset coming in $16*32$ size.

$\bullet$ \texttt{crates}: A puzzle game inspired by Sokoban \citep{GAME_sokoban} in which the player has to push crates around the level into slots. The player can only push crates, not pull them, and the crates can only be pushed one at a time. Although seemingly easy, game levels can be hard to play as even a single push of crates into constrained positions, such as against walls, corners, or narrow passages, can make it immovable or block access to other areas. There are $10,000$ levels in this dataset coming in $16*16$ size and 2 crates.

$\bullet$ \texttt{vertical}: A platformer game inspired by Super Cat Tales \citep{GAME_supercat}. This game is interesting in that the player does not jump directly from a standstill, but can wall climb, jump off walls, and leap off ledges. Levels generally proceed vertically, where each level begins on the bottom platforms with a farm-like atmosphere. As the player goes up through the level, the platforms become increasingly industrialized. There are more than $10,000$ levels in this dataset coming in $20*16$ size.

\begin{figure*}[h]
    \includegraphics[width=\textwidth]{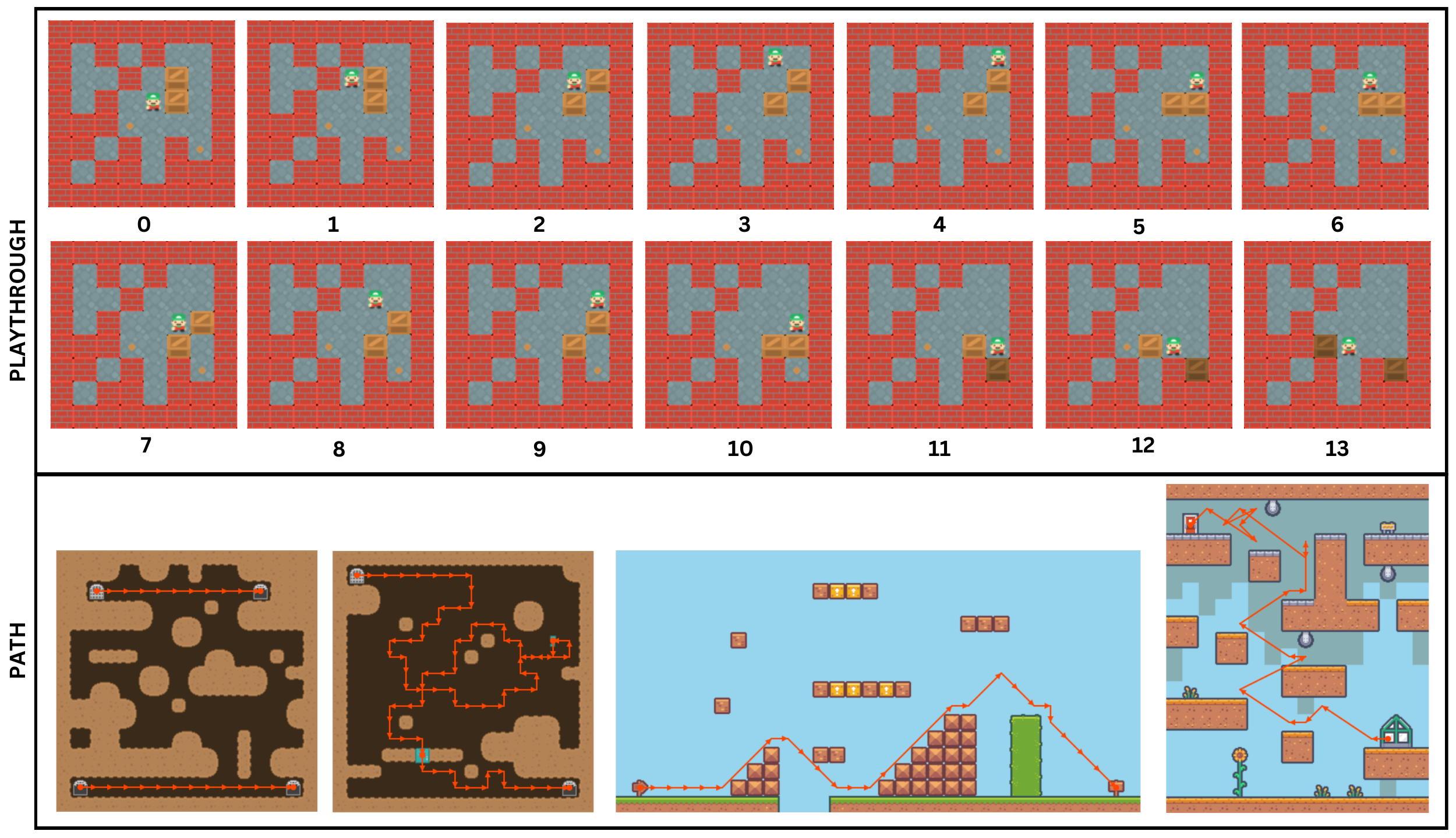}
    \caption{Level solution examples. Some solutions come in the format of the player's path in the level, and some (\texttt{crates}) come in the format of step by step playthrough.}
    \label{fig:solutions}
\end{figure*}

Various techniques have been employed to enforce global constraints within game levels through the integration of player path representations \citep{sarkar2020exploring,summerville2016super}. \citet{Summerville_Guzdial_Mateas_Riedl_2021} trained an LSTM model on Super Mario Bros. levels by incorporating the representations of player paths into the game level. Subsequently, \citet{sorochan2021generating} attempted to enhance controllability and coherence by leveraging models trained on Lode Runner levels with the played path by human players. Therefore, we included the solutions of all levels as part of the metadata in this dataset as it could be useful for downstream applications. Samples of such game solutions are illustrated in Figure \ref{fig:solutions}.
\begin{table*}[h]
    \centering
    \begin{tabularx}{\textwidth} { 
       >{\centering\arraybackslash}p{0.12\textwidth}
      | >{\centering\arraybackslash}p{0.838\textwidth} }
     \hline
      Dataset & Tiles\\
     \hline
      \texttt{cave}  & \includegraphics[height=0.5cm]{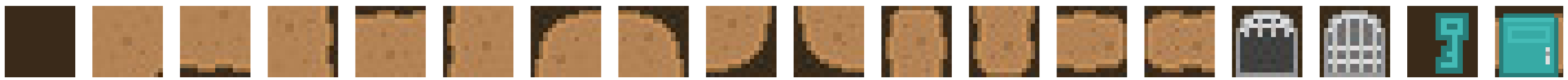} \\
     \hline
     \texttt{platform}  & \includegraphics[height=0.5cm]{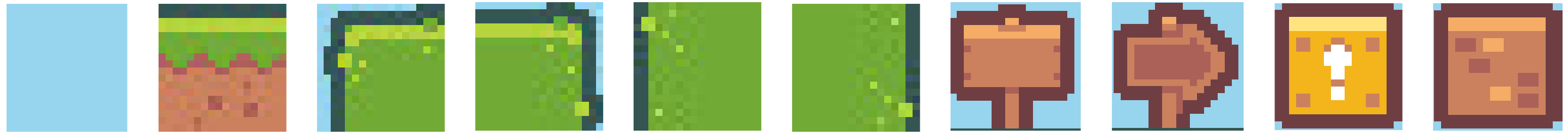} \\
     \hline
     \texttt{crates}  & \includegraphics[height=0.5cm]{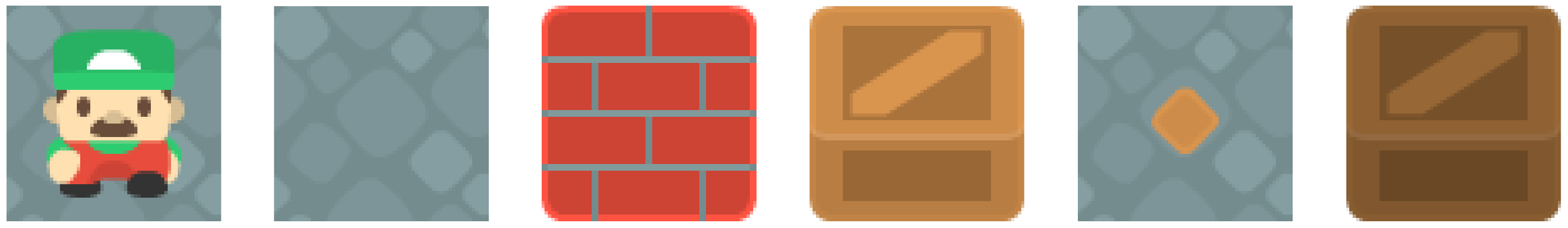} \\
     \hline
     \texttt{vertical}  & \includegraphics[height=0.5cm]{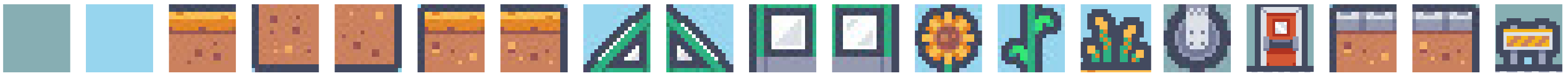} \\
     \hline
    \end{tabularx}
    \caption{Tile types available in each dataset.}
    \label{table:tiles}
\end{table*}

\section{Experiments} \label{sections:experiments}
In this section, we conduct experiments to showcase the unique attributes of this dataset, specifically the hard constraints these game levels inherently include that make them sensitive to small input changes.

\subsection{Data Collection} \label{sections:collection}
In this work we generated levels using the Sturgeon \citep{cooper2022sturgeon} constraint-based level generator. Sturgeon is a procedural content generation system designed for tile-based games that relies on constraints. Sturgeon is open source and the code is available on GitHub\footnote{\url{https://github.com/crowdgames/sturgeon-pub}}. Sturgeon utilizes a concise, mid-level API to define constraints on Boolean variables, converting these into low-level constraint satisfaction problems. These problems can then be solved using various standard low-level solvers, such as SAT, SMT, or Answer Set solvers.

Sturgeon itself can incorporate local tile patterns as constraints, ensuring that generated levels contain the same local structures as example levels.  In this work, we used a few human-designed example levels of ours to provide these patterns, and then generated a much larger number of levels based on these examples. It can also incorporate constraints on overall tile counts, so that the general makeup of generated levels is similar to the examples.

To ensure solvability, Sturgeon can use two different approaches.  First, Sturgeon incorporates pathfinding constraints between specific tiles; most commonly used to ensure there is a navigable path from the start to the goal. These constraints can also be used, for example, to ensure that keys and doors are reachable. These pathfinding constraints are based on a \emph{reachability template} that represents how the player can move and can vary from game to game. Sturgeon outputs the path found with the solution as a list of location-to-location edges. Because this path can often be circuitous, a more direct path can also be computed in a post-processing step. The second approach to ensuring solvability is more involved, and represents game mechanics as tile rewrite rules \citep{cooper_sturgeonmkiii_2023}. With this approach, in addition to the level, an entire playthrough is generated.  The playthrough is a sequence of game states (which are themselves essentially levels), where the change from one state to the next is constrained by the tile rewrite rules.  The level must be completed by the ending state of the playthrough.

The unsolvable levels are generated using Sturgeon's unreachability constraint \citep{cooper_literally_2024} that uses constraints that the level's goal is not reachable from its start. Unsolvable \texttt{crates} levels are created by automatically modifying a solvable level to move a crate to a non-valid location (like corners), and removing or adding crates and slots.

Sturgeon uses an approach that involves shuffling variable ordering to introduce randomness in solutions. While this method does not guarantee the creation of entirely unique levels, our preliminary analysis of the generated levels revealed no major issues with duplicate levels in datasets.

Each level contains two representations: an image and a text description. The image representation is a picture of the level and the text representation is a 2D tile of characters, where each character generally indicates the function of the tile at that tile location (e.g., typically \texttt{X} for solid and \texttt{-} for empty). These representations vary from game to game. Also, the same text representation can correspond to different images (same functionality, different aesthetics).

Each level, if solvable, also has an associated solution. This is either a location-to-location edge path from the start location (usually represented by a \texttt{\{}) to the goal location (usually represented by a \texttt{\}}), or a playthrough sequence of levels.  The solutions are not necessarily unique and there may be (many) other solutions to a given level.

In this work, levels were generated by a portfolio of low-level constraint solvers. Solvers used were PySAT's RC2 \citep{ignatiev_pysat_2018,ignatiev_rc2_2019} solver without and with builtin cardinality constraints (\citep{liffiton_cardinality_2012}), clingo's backend Answer Set solver \cite{gebser_potassco_2011}, and SciPy's MILP solver \citep{virtanen_scipy_2020} based on HiGHS \citep{huangfu_parallelizing_2018}.  The portfolio runs these solvers in parallel, takes the solution from the first to finish, and then stops the other solvers.

\subsection{Measuring Non-robustness}
The non-robustness metric can be calculated in both discrete and continuous forms. In the continuous form, the choice of radius $r$ may need adjustment to account for relative scaling. Using the law of total expectation, we first sample uniformly at random and then compute the expected proportion of data points within a radius $r$ in Euclidean distance that have dissimilar labels compared to the sampled data.

In the discrete form, a similar process is followed, but the radius $r$ is fixed at 1, representing the nearest neighbors in Hamming distance. We then calculate the expectation of the proportion of data points with dissimilar labels.

We use the discrete form to calculate the non-robustness of the true distribution, while the continuous form is applied to evaluate the non-robustness of the generated dataset. This allows for a comparison with state-of-the-art machine learning datasets.

\section{Results}
\subsection{Distribution Non-robustness}
In this section, we assess the non-robustness of data distribution by generating $100$ random levels from each game’s distribution. These levels are generated with the exact same process the GGLC levels are generated. For each level, we explore all possible radius-1 changes (measuring non-robustness in discrete form), where a single tile is randomly replaced by another tile (only one random replacement, not swaps). We then analyze how these changes affect the level's solvability and acceptability. The table below presents the percentage of single-tile changes that altered the solvability of the levels. For the game levels in our dataset, $100$ levels amount to at least 16 \time 16 changes per level. We determined that this sample size was sufficient to illustrate the sensitivity of the levels to distributional changes. 

Table \ref{tab:true-non-robustness} highlights the high sensitivity of game levels to small input changes. Different games exhibit varying degrees of sensitivity, influenced by their gameplay mechanics and level design structures. For example, \texttt{crates} shows the highest sensitivity, where small changes, such as moving a crate into a corner, often result in unsolvable levels due to its mechanics' reliance on precise crate placement. On the other hand, \texttt{vertical} shows high sensitivity in terms of acceptability, as its visual design features multiple tile types with the same functionality but distinct decorative appearances. \texttt{cave} shows sensitivity to input change that both results in unsolvability and unacceptability. This is inherited from the rigid framework of mazes and the versatility of tiles in this game.


\begin{table*}[h]
\centering
\begin{tabularx}{\textwidth}{ 
       >{\centering\arraybackslash}p{0.25\textwidth}
      | >{\centering\arraybackslash}p{0.15\textwidth}
      | >{\centering\arraybackslash}p{0.15\textwidth}
      | >{\centering\arraybackslash}p{0.15\textwidth}
      | >{\centering\arraybackslash}p{0.15\textwidth}
      }
\hline
& \texttt{cave} & \texttt{platform} & \texttt{crates} & \texttt{vertical} \\
\hline
Changed Solvability & 43.1\%  & 17.1\% & 78.9\% & 17.5\%\\
Changed Acceptability & 43.0\% & 0.1\% & 0\% & 42.5\% \\
\hline
\end{tabularx}
\caption{Non-robustness of distribution}
\label{tab:true-non-robustness}
\end{table*}

\begin{figure*}[ht]
    \centering
    \begin{subfigure}[b]{0.24\textwidth}
        \centering
        \includegraphics[width=\textwidth]{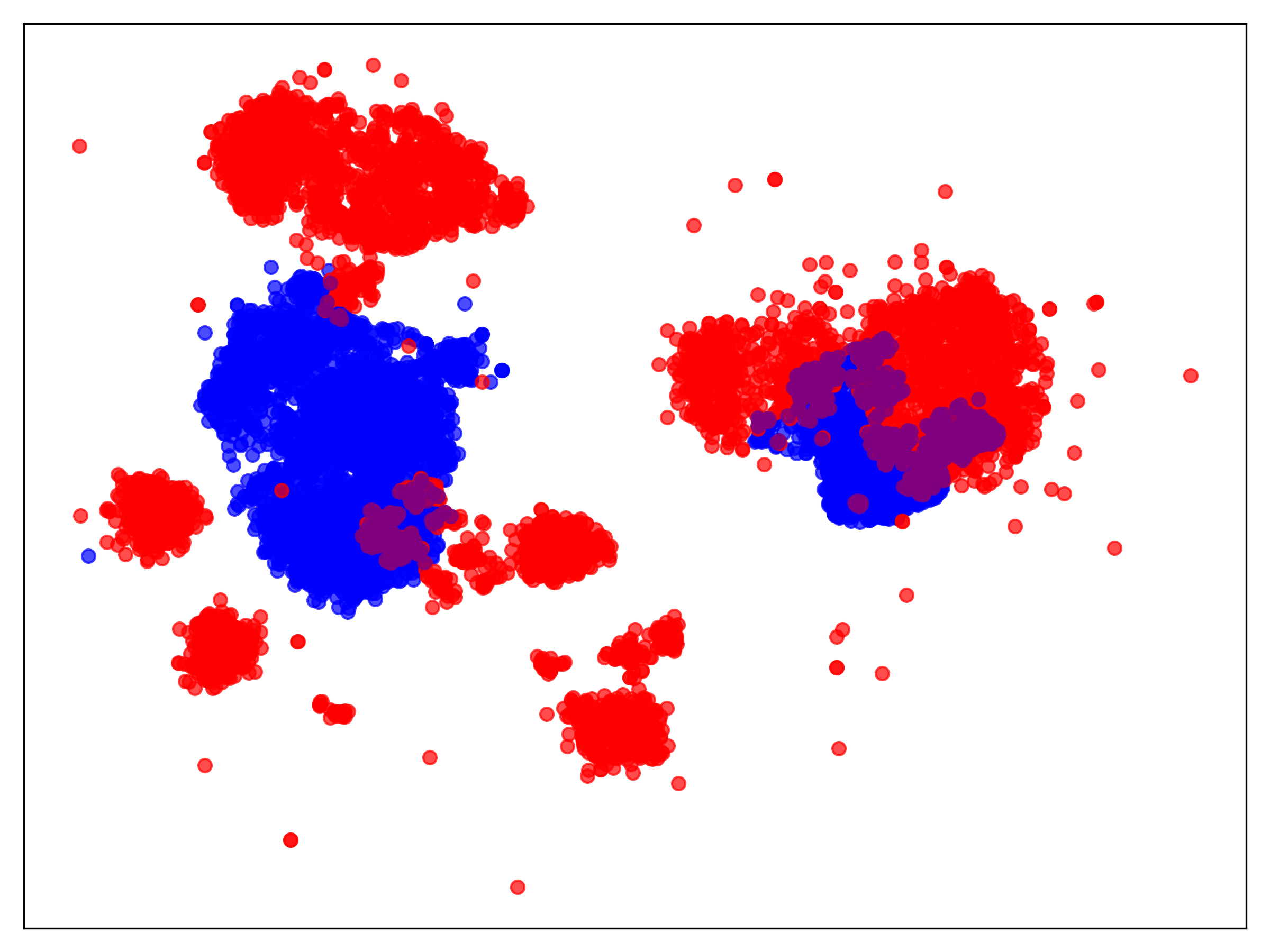}
        \caption{\texttt{cave}}
    \end{subfigure}
    \begin{subfigure}[b]{0.24\textwidth}
        \centering
        \includegraphics[width=\textwidth]{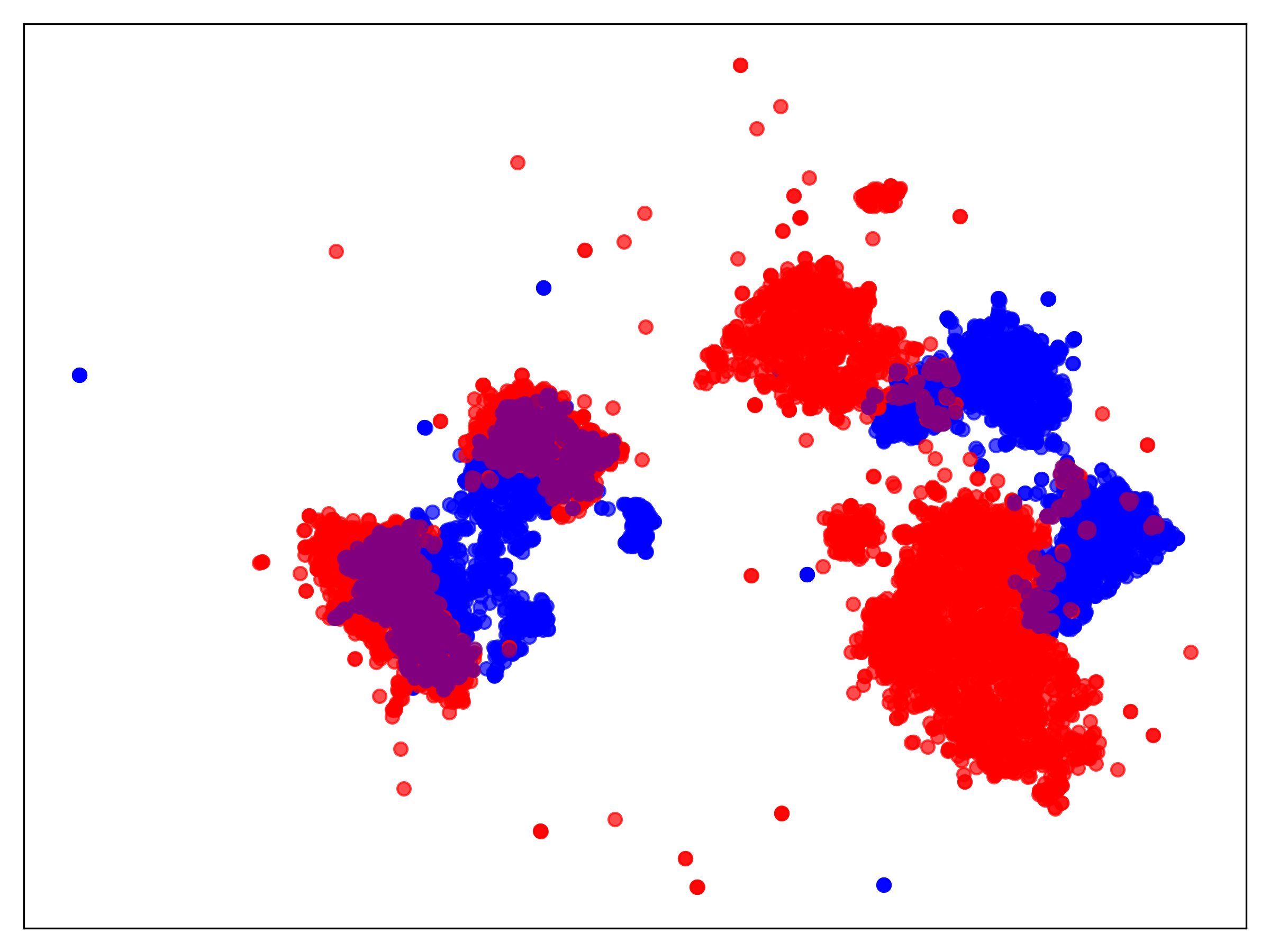}
        \caption{\texttt{platform}}
    \end{subfigure}
    \begin{subfigure}[b]{0.24\textwidth}
        \centering
        \includegraphics[width=\textwidth]{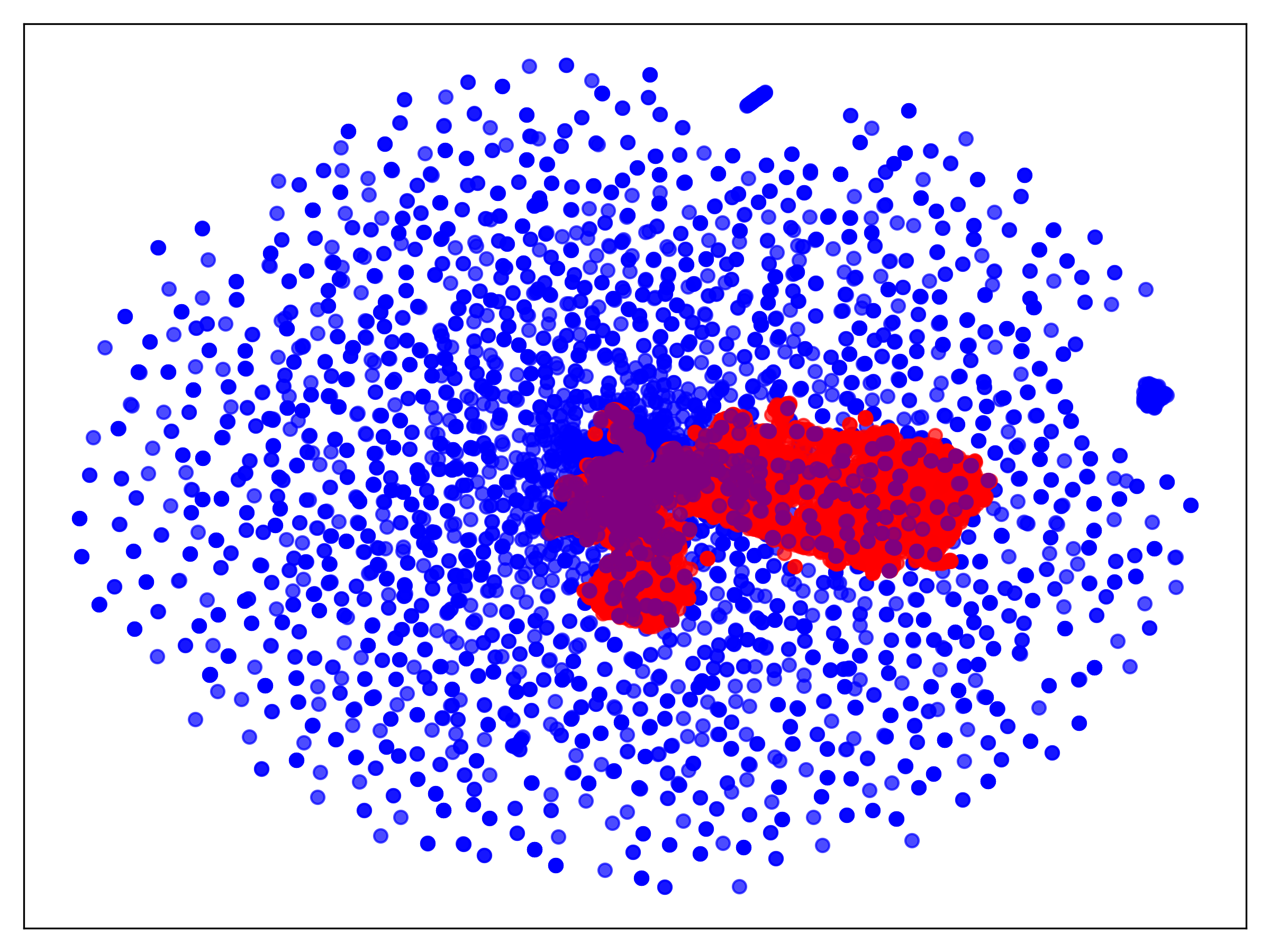}
        \caption{\texttt{crates}}
    \end{subfigure}
    \begin{subfigure}[b]{0.24\textwidth}
        \centering
        \includegraphics[width=\textwidth]{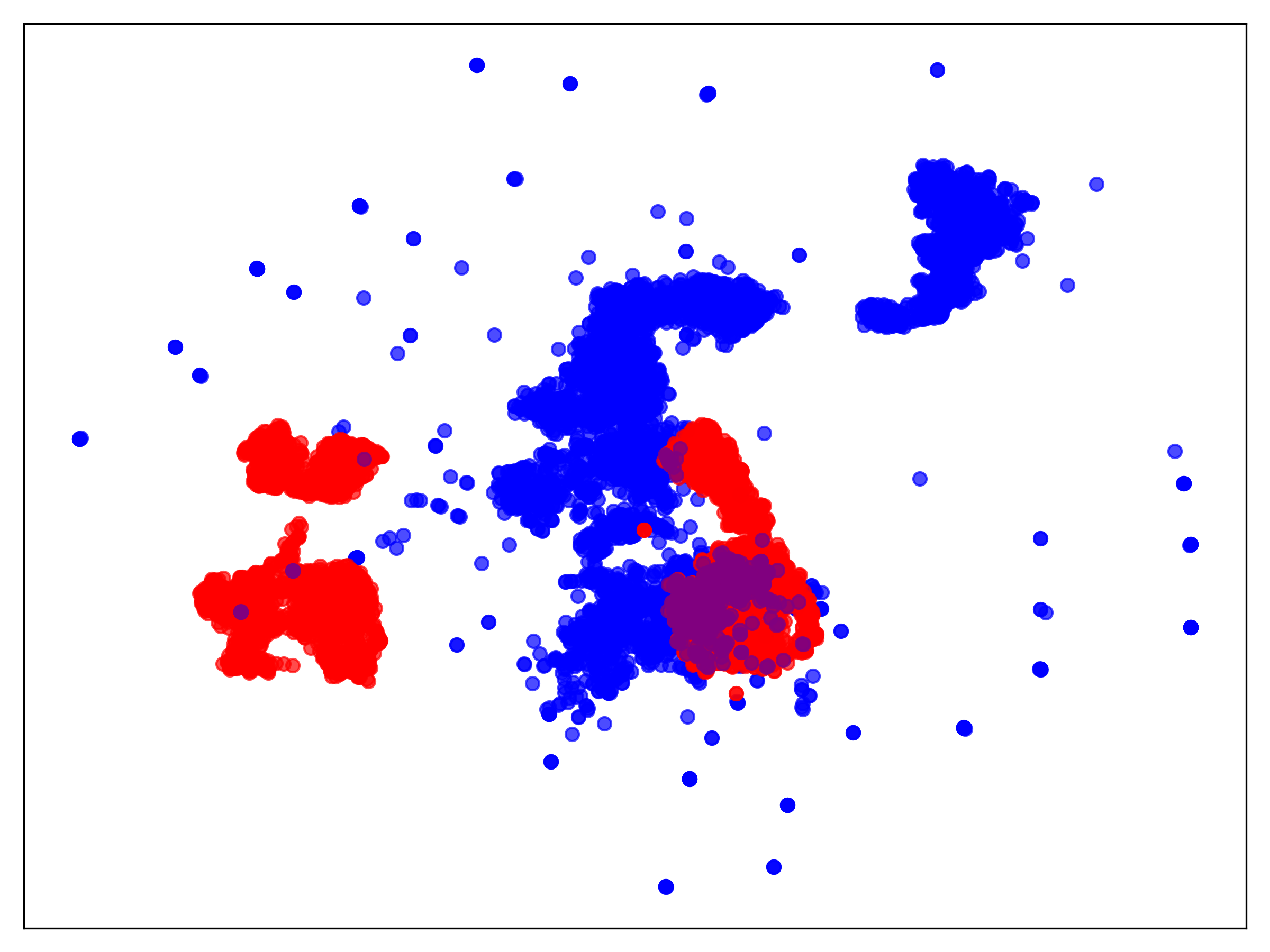}
        \caption{\texttt{vertical}}
    \end{subfigure}
    \caption{UMAP visualization of CLIP embedding space with blue for solvable levels and red for unsolvable levels.  The purplish color appears where the two classes, solvable and unsolvable, overlap.}
    \label{fig:distribution}
    
\end{figure*}

\begin{table*}[h]
    \centering
    \begin{tabularx}{1\textwidth} { 
      >{\centering\arraybackslash}p{1.5cm}
      *{10}{>{\centering\arraybackslash}X} }
     \toprule
       & \multicolumn{10}{c}{Radius} \\
      \cmidrule(lr){2-11}
      Dataset & $0.00001$ & $0.00005$& $0.0001$& $0.0005$& $0.001$& $0.005$& $0.01$& $0.05$& $0.1$ & $avg$\\
     \midrule
     \texttt{CIFAR-10} & 0.0 & 0.0 & 0.1 & 2.9 & 6.7 & 12.4 & 12.8 & 14.4 & 18.9 & 7.6 \\
    \texttt{MNIST} & 0.0 & 0.0 & 0.1 & 1.6 & 3.7 & 5.8 & 6.2 & 6.1 & 10.1 & 3.7 \\
     \texttt{cave} & 0.0 & 0.7 & 2.3 & 24.3 & 29.6 & 32.6 & 32.6 & 35.2 & 36.2 & 21.5 \\
     \texttt{platform} & 0.0 & 0.0 & 0.0 & 2.0 & 5.9 & 12.5 & 14.1 & 21.2 & 28.1 & 9.3 \\
    \texttt{crates} & 0.0 & 0.0 & 0.3 & 2.5 & 6.0 & 13.1 & 15.3 & 20.8 & 25.8 & 9.3 \\ 
    \texttt{vertical} & 0.0 & 0.0 & 0.1 & 1.1 & 3.2 & 5.6 & 6.7 & 10.3 & 15.9 & 4.8 \\
     \bottomrule
    \end{tabularx}
\caption{Non-robustness of samples. A higher number indicates more sensitivity of data to changes in the input.}
\label{table:metric}
\end{table*}
\subsection{Sample Non-robustness}
As discussed in earlier sections, we generated a synthetic dataset containing both solvable and unsolvable game levels. This dataset not only serves as a resource for PCGML downstream applications but also allows us to illustrate our problem statement using real-world data. We need to use an embedding space as a surrogate to be able to compare the sensitivity of input changes in our dataset with a categorical nature to the common non-structured machine learning datasets such as CIFAR-10. Since our dataset includes both images of the game levels as well, we will use the images to perform the same embedding and dimensionality reduction between all datasets. This will enable us to compare them more effectively (measuring non-robustness in continuous form).

We used \citet{berns2024not} study on the alignment of similarity metrics with human judgment on tile-based video game levels to choose our embedding. They studied seven similarity estimation metrics to determine the most accurate metric for representing human expectations regarding the similarity of tile-based game levels and their findings indicate that CLIP \citep{radford2021learning} exhibited one of the highest overall agreement between the metrics and human judgment. Therefore, we utilized the CLIP model from the Hugging Face Transformers library \citep{wolf-etal-2020-transformers} to embed the levels.

After embedding the data using CLIP, we employed Uniform Manifold Approximation and Projection (UMAP) for dimension reduction to a 2D space that we can measure Euclidean distance effectively. This technique was chosen because our preliminary studies indicated that it provided the best results for distinguishing between game level classes and facilitating visualization. Figure  \ref{fig:distribution} illustrates the distributions visualized through this method. A visual inspection of the distributions reveals that in some game datasets, there is a big overlapping between solvable and unsolvable classes.

To compare the non-robustness of the data between all datasets, we normalize each data to a common range $[0, 1]$ using the minimum and maximum values.  Since the purpose of this metric is to evaluate the sensitivity of the data to very small changes, we calculated non-robustness for the sample data in normalized distribution for radius $r$ ranging from the smallest value ($0.00001$) with $0$ value for all datasets and increasing the $r$ to $0.1$.

The analysis of the non-robustness of data (relative to unsolvability) within each dataset is shown in Table \ref{table:metric}. The results are compatible with the expected results from the true distribution of non-robustness which indicates that the created dataset is a good representation of the true problem statement. Within this dataset, \texttt{cave} (representative of maze-like games) and \texttt{crates} (representative of puzzle games) demonstrate the highest level of non-robustness among the datasets (more sensitive to changes).

It should be noted that, for the computation of this metric, we employ the optimal embeddings for tile-based games (CLIP) \citep{berns2024not}, which may not be the most suitable for baseline datasets (MNIST and CIFAR10) leading to inflated non-robustness values. Nevertheless, even with these inflated values --- and despite the fact that a single tile change corresponds to 16~$\times$~tile pixel change --- some game datasets still exhibit greater sensitivity to small changes compared to MNIST and CIFAR-10.


    
    


\section{Discussion}
The GGLC dataset includes levels inspired by common genres, that are not directly sourced from existing games. This contrasts with datasets like the VGLC, which is taken from well-known and popular games. This comes with the advantage of having an expansive dataset of game levels specifically made for the purpose of research. However, this means that these games are fairly simple and short and currently are not created to be played as they lack major gameplay aspects like moving elements, enemies, and items or powerups to collect. Nevertheless, this could also be seen as a flexibility that allows for diverse interpretations and potential innovations in gameplay design.

\section{Conclusion}
This study aims to formalize a phenomenon observed in tile-based game levels --- an example of highly structured discrete data --- where changes to individual tiles can significantly affect the overall level, potentially rendering it unsolvable or undesirable.  To quantify this sensitivity to input changes across datasets, we introduce the concept of non-robustness of data as a metric. Furthermore, we present a comprehensive dataset of 2D tile-based games, available under a CC-BY 4.0 license. Our work highlights the differences between these game levels and commonly used datasets in the machine learning community. We hope this dataset will help the challenges associated with data sparsity and other issues in procedural content generation and bridge different communities interested in confronting the unique challenges posed by novel datasets.

\begin{acks}
Support was provided by Research Computing at Northeastern University (\url{https://rc.northeastern.edu/}) through the use of the Discovery Cluster.
\end{acks}

\bibliographystyle{ACM-Reference-Format}
\bibliography{refs}

\end{document}